\pdfoutput=1

\documentclass[11pt]{article}

\usepackage[final]{acl}

\usepackage{times}
\usepackage{latexsym}

\usepackage[T1]{fontenc}

\usepackage[utf8]{inputenc}

\usepackage{microtype}

\usepackage{inconsolata}

\usepackage{graphicx}

\title{Improving Language Models for Emotion Analysis: Insights from Cognitive Science}

\author{
  Constant Bonard \\
  University of Bern, Institute of Philosophy, Hochschulstrasse 4, 3012 Berne, Suisse \\
  \texttt{constant.bonard@gmail.com}
  \AND
  Gustave Cortal \\
  Université Paris-Saclay, ENS Paris-Saclay, CNRS, LMF, 91190, Gif-sur-Yvette, France \\
  Université Paris-Saclay, CNRS, LISN, 91400, Orsay, France \\
  \texttt{gcortal@ens-paris-saclay.fr} \\
}

\begin{document}
\maketitle
\begin{abstract}
We propose leveraging cognitive science research on emotions and communication to improve language models for emotion analysis. First, we present the main emotion theories in psychology and cognitive science. Then, we introduce the main methods of emotion annotation in natural language processing and their connections to psychological theories. We also present the two main types of analyses of emotional communication in cognitive pragmatics. Finally, based on the cognitive science research presented, we propose directions for improving language models for emotion analysis. We suggest that these research efforts pave the way for constructing new annotation schemes, methods, and a possible benchmark for emotional understanding, considering different facets of human emotion and communication.
\end{abstract}

\def\thefootnote{*}\footnotetext{The authors contributed equally and are listed in alphabetical order.}

\section{Introduction}
Emotion analysis in natural language processing aims to develop computational models capable of discerning human emotions in text. Recently, language models have been widely used to solve various tasks in natural language processing, including emotion analysis \cite{devlin-etal-2019-bert,brownLanguageModelsAre2020a}. This field of research faces several limitations. First, different ways of conceptualizing emotions lead to different annotation schemes and datasets \cite{klingerWhereAreWe2023}. As a result, the generalization ability of models is limited, and it is often impossible to compare studies. To address these limitations, it has been proposed to unify some annotation schemes based on the semantic proximity of emotion categories \cite{bostan-klinger-2018-analysis}, to automatically find emotion categories from data \cite{de-bruyne-etal-2020-emotional}, or to obtain emotion embeddings independent of annotation schemes \cite{buechel-etal-2021-towards}. Inspired by psychology and cognitive science research, we believe building an annotation scheme unifying different perspectives on the emotional phenomenon would be possible and desirable.

In addition, existing benchmarks evaluate certain aspects of emotional understanding but do not consider its full complexity \cite{campagnano-etal-2022-srl4e,zhangSentimentAnalysisEra2023,paechEQBenchEmotionalIntelligence2024}. For example, \citet{paechEQBenchEmotionalIntelligence2024} proposes to evaluate the emotional understanding of language models by predicting the intensity of emotions in conflict scenes. This type of evaluation is too limited: benchmarks should reflect as much as possible the richness of emotional understanding in humans, a richness documented in different branches of affective sciences \cite{green_self-expression_2007,wharton_that_2016,scarantino_how_2017,barrett_emotional_2019,bonard_emotion_2023}.

Another related research area focuses on the theory of mind of language models, \textit{i.e.}, their ability to correctly attribute mental states to others. In our view, this literature is promising in that it links recent developments in language models to theories and empirical methods in cognitive science (for a review, see \citet[section 5]{bonard_can_2024}). Notably, several tasks and benchmarks have been developed to measure the ability of language models to succeed at different versions of the False Belief Task \cite{trott_large_2022,aru_mind_2023,gandhiUnderstandingSocialReasoning2023,holterman_does_2023,kosinski_theory_2023,mitchell_debate_2023,shapira_clever_2023,stojnic_commonsense_2023,ullman_large_2023}. However, theory of mind and, more generally, social reasoning abilities go beyond the ability to succeed at the False Belief Task \cite{apperly_humans_2009,langley_theory_2022, maHolisticLandscapeSituated2023}. The ability to correctly interpret expressed emotions cannot be reduced to it. The degree to which language models possess this emotional competence is worth studying in its own right.

Generally speaking, research on language models for emotion analysis would benefit from cognitive science research on emotion and communication. In particular, we believe this approach can lead to better ways of annotating emotions expressed in text. Additionally, it can improve the evaluation of the emotional understanding of language models by developing new benchmarks. In what follows, we present an overview of psychological theories of emotion (section \ref{theorie_emotion}) and ways of annotating emotions in natural language processing (section \ref{analyse_des_emotions}). Then, inspired by specific psychological and linguistic theories (section \ref{modele_detective}), we propose research directions to address some of the current limitations of emotion analysis (section \ref{directions}).

\paragraph{Contributions.} We propose integrating different cognitive science theories on emotion with NLP research. We explain why and how emotion analysis should use research from cognitive pragmatics, specifically what we call "the detective analysis", to improve automatic emotion analysis. We suggest that these points lead both to devising a new annotation scheme and improving how language models should be evaluated for emotion analysis.

\section{Emotion Theories in Cognitive Science\label{theorie_emotion}}

This section will present the three main emotion theories in psychology to provide a background for connecting emotion analysis in natural language processing with cognitive science.
\paragraph{Basic emotion theory.} Basic emotion theory is certainly the most influential today. Inspired by Darwin’s research on emotions \cite{darwin_expression_1872}, it postulates a certain number of discrete, basic emotions that are universal and innate among humans due to their evolutionary origins. Emotions are understood as psycho-physiological "programs" that were naturally selected to help overcome recurrent evolutionary challenges \cite{cosmides_evolutionary_2000}. A prominent version is that of Paul Ekman \cite{ekman_basic_1999}, who sought to show, as Darwin envisaged, that some emotions are expressed with the same facial expressions across cultures – Ekman used Darwin’s \cite{darwin_expression_1872} list of six "core" expressions of emotions: anger, fear, surprise, disgust, happiness, and sadness. He notably conducted studies with individuals having no exposure to Western culture, indicating that they could accurately identify facial expressions for these six emotions \cite{ekman1971constantsac}. It should be noted that Ekman left it open how many basic emotions there are. Besides the six emotions listed, candidates include amusement, contempt, embarrassment, guilt, pride, and shame \cite{ekman_basic_1999}. Other versions of basic emotion theory have different lists \cite{tomkins_affect_1962,izard_basic_1992,panksepp_affective_1998,plutchikNatureEmotionsHuman2001}. For a discussion of the evidence supportive of basic emotion theory, notably the potential physiological and neurological signatures of basic emotions, see \citet[129––131]{moors_demystifying_2022}.

\paragraph{Psychological constructivism.} Psychological constructivism is the most influential alternative to basic emotion theory today. It rejects that there are discrete, basic emotions universally shared by humans and posits instead that emotion kinds such as anger, fear, and joy are constructed through the interplay of biological, psychological, and sociocultural factors. Early proponents include \citet{schachter_cognitive_1962}, but its main representatives are James Russell and Lisa Feldman Barrett \cite{russell_core_1999}. Psychological constructivists focus on the feeling component of emotions that they interpret as a continuum with no categorical barriers. Feelings are typically represented in a two-dimensional space with a valence axis (pleasant–unpleasant feelings) and an arousal axis (feelings of activation–deactivation). The impression that there are discrete emotions is seen as a social construct: different forms of enculturation yield different ways to conceptualize or label our bodily feelings into discrete emotional kinds. For a discussion of the evidence supportive of psychological constructivism, see \citet[261––265]{moors_demystifying_2022}. Some evidence comes from so-called "arousal misattribution" studies, i.e. cases where subjects misinterpret the source of their arousal and where that seems to influence what emotions they undergo.

\paragraph{Appraisal theory.} The third major psychological theory of emotion is appraisal theory, whose empirical version was pioneered by Magda Arnold \cite{arnold_emotion_1960}. It was developed to explain the absence of a bijective, one–to–one correspondence between kinds of emotions and emotional stimuli, \textit{i.e.}, the fact that the same kind of stimuli triggers different emotions and that different kinds of stimuli trigger the same kind of emotion. To explain this fact, appraisals are postulated as mediators between stimuli and emotional reactions. Appraisals are cognitive evaluations (unconscious, fast, and error-prone) of the relevance of stimuli given one’s concerns and how one should react. Appraisal theory hypothesizes that, for instance, Sam is fearful of the mouse in the kitchen because he appraises it as an imminent threat to his safety, while Maria, on the other, is angry that there is a mouse in the kitchen because she appraises it as an intruder to be kicked out. Thus, each emotion kind can be analyzed by the associated appraisal. For instance, \citet{lazarus_progress_1991} proposes \textit{imminent danger} for fear, \textit{demeaning offense} for anger, \textit{irrevocable loss} for sadness, and \textit{progress towards a goal} for happiness. 

In the 1980s, appraisal theorists started to analyze appraisals as regions in a multi-dimensional space \cite{moors_appraisal_2013}. Appraisal dimensions typically include (a) the goal-conduciveness of the stimulus, (b) the coping potential of the individual in the situation, (c) the urgency of the needed response, (d) the cause of the eliciting event (me, others, intentional or not), and (e) the compatibility with one’s normative standards. For instance, fear is triggered by an appraisal of a stimulus as (a) highly inconducive, (b) hard to cope with, and (c) requiring an urgent response.
For a discussion of the evidence supportive of appraisal theory, see \citet[190––196]{moors_demystifying_2022}. Most evidence comes from self-report studies where participants are asked to recall instances of emotions and to rate these in terms of appraisal variables. Other evidence comes from manipulating appraisal dimensions and measuring associated emotions (\textit{e.g.}, in a video-game setting) or from neurological predictions about correlations between brain activations and appraisal dimensions.

\begin{figure}[!htb]
    \centering
    \includegraphics[scale=0.19]{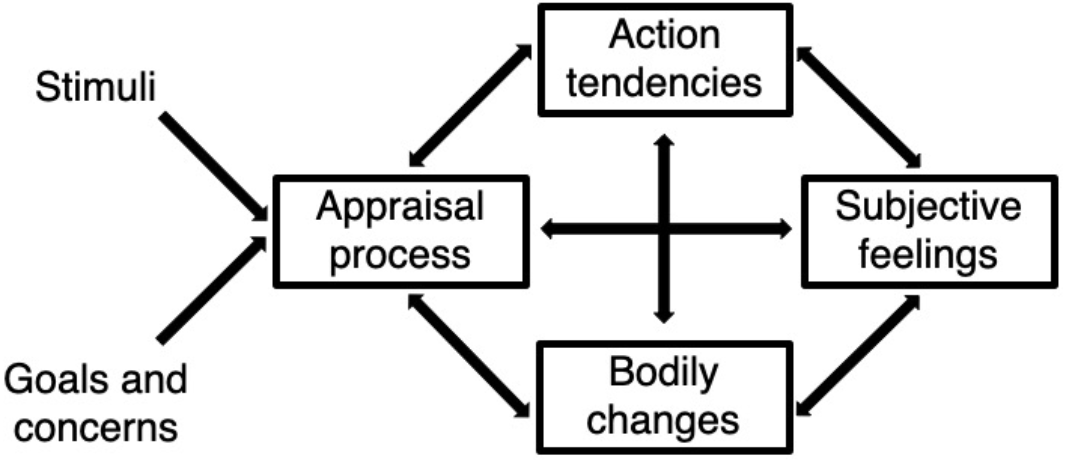}
    \caption{The integrated framework for emotion theories. Rectangles represent the four components constituting an emotional episode, and arrows represent causation. Adapted from \citet{scherer_emotion_2019}.}
    \label{fig:scherer}
\end{figure}

\paragraph{An integrated framework for emotion theories.} Though the three theories reviewed are usually considered rivals, some have argued for their integration \cite{scherer_emotion_2019,bonard_emotions_2021,schererTheoryConvergenceEmotion2022}. Arguably, the three theories differ mainly in their focus. Basic emotion theory focuses on the universal traits inherited from evolution, particularly their physiological and bodily expressions. Psychological constructivism focuses on the dimensions of feeling and how individuals categorize them. Appraisal theory focuses on emotional elicitation and action tendencies. We believe that a framework integrating the various elements studied by these theories is possible and desirable. What we call "the integrated framework for emotion theories" proposes to do so by postulating that paradigmatic emotional episodes are made of synchronized and causally interconnected changes in four components: appraisal process, action tendencies, bodily changes (motor expressions and physiological responses), and subjective feelings. For a discussion of this integrated framework, see \citet{schererTheoryConvergenceEmotion2022}. 

\section{Emotion Analysis in Text\label{analyse_des_emotions}}

\subsection{How is emotion annotated in text?\label{comment_emotion_annotee}}

\paragraph{Emotion is a category.} Textual emotion analysis relies on basic emotion theories to define different emotion categories to associate with textual units (a textual span, a sentence, or a document). For instance, the sentence "I love philosophy." could automatically be associated with the discrete emotion \textit{happiness}. Several annotation schemes focus on subsets of categories while others encompass a broader set, reaching over 28 different categories \cite{demszky-etal-2020-goemotions,bostan-klinger-2018-analysis}.

\paragraph{Emotion is a continuous value with affective meaning.} Instead of representing emotion as a category, some annotation schemes consider emotion as a point in a multidimensional space, associating continuous values with textual units \citep{buechel-hahn-2017-emobank}. These dimensions carry an affective meaning. Two dimensions dominate the literature and stem from psychological constructivism, which considers, as we have seen, that an emotion can be characterized by its degree of \textit{pleasantness} and its degree of \textit{arousal}. Thus, the sentence "His voice soothes me." could be automatically associated with two continuous values: a degree of \textit{pleasantness} of 4 out of 5 and a degree of \textit{arousal} of 1 out of 5.

\paragraph{Emotion is a continuous value with cognitive meaning.} These dimensions can also carry a cognitive meaning. Recently, a new line of research proposes incorporating appraisal theories into emotion analysis models \cite{hofmann-etal-2020-appraisal,troianoDimensionalModelingEmotions2022,zhanEvaluatingSubjectiveCognitive2023}. From this perspective, emotions are caused by events evaluated according to several cognitive dimensions. For example, the sentence "I received a surprise gift." could be automatically associated with several continuous values: the event is \textit{sudden} (4 out of 5), \textit{contrary to social norms} (0 out of 5), and the person has \textit{control} over the event (0 out of 5).

\paragraph{Emotion consists of semantic roles.} An emotion cannot be reduced to a category or continuous values with affective or cognitive meaning. To better understand an emotional event, several approaches associate spans of text with semantic roles, such as \textit{cause}, \textit{target}, \textit{experiencer}, and \textit{cue} of the emotion \cite{lee-etal-2010-text,kimWhoFeelsWhat2018,bostan-etal-2020-goodnewseveryone,oberlanderExperiencersStimuliTargets2020,campagnano-etal-2022-srl4e,weggeExperiencerSpecificEmotionAppraisal2023,cortal-2024-sequence-sequence}. Thus, instead of considering emotion as caused by an event, semantic role labeling of emotions considers that emotion \textit{is} an event \cite{klingerWhereAreWe2023} that must be reconstructed by answering the question: "Who (\textit{experiencer}) feels what (\textit{cue}) towards whom (\textit{target}) and why (\textit{cause})?". In this example, each text span can be associated with a semantic role: "Louise (\textit{experiencer}) was angry (\textit{cue}) at Paul (\textit{target}) because he did not warn her (\textit{cause})."

\paragraph{Emotion is a refined feeling.} Sentiment analysis, a fundamental task in natural language processing, is sometimes considered a simplified version of emotion analysis. In its most basic form, sentiment analysis associates textual units with a category indicating a polarity (\textit{positive} or \textit{negative}) \cite{poriaTipIcebergCurrent2020a}. A finer-grained task identifies aspects of a product or topic and determines the sentiment expressed about each of these aspects \cite{zhangSurveyAspectBasedSentiment2022}. For example, in the sentence "The battery life of this phone is amazing, but its camera quality is disappointing.", the sentiment is \textit{positive} for the aspect "battery life" and is \textit{negative} for the aspect "camera quality."

\subsection{Limitations\label{limitations}}

\paragraph{No unified annotation scheme.} Divergences in the psychological definition of emotion lead to divergences in how emotion is annotated in the text. Psychological theories of emotions represent different perspectives on the emotional phenomenon. However, these perspectives are not as contradictory as they seem and may even tend towards unification (section \ref{theorie_emotion}). We believe this is also the case for annotation schemes in emotion analysis. In section \ref{directions}, we provide directions for constructing a unified annotation scheme inspired by recent debates in psychology \cite{schererTheoryConvergenceEmotion2022}.

\paragraph{Emotion verbalization is overlooked.} Emotion analysis rarely considers the process of emotion verbalization. As a result, it is difficult to obtain annotation guides that clearly define the linguistic markers to annotate in text. We want to highlight the linguistic theory of Raphael Micheli, which categorizes a broad panel of linguistic markers into three emotion expression modes \cite{Micheli2014}: \textit{labeled}, \textit{displayed}, and \textit{suggested} emotion. Emotion can be expressed explicitly with an emotional label ("I am \textit{happy} today"), be displayed with linguistic characteristics of an utterance such as interjections and punctuations ("\textit{Ah!} That's great \textit{!}"), or be suggested with the description of a situation that, in a given sociocultural context, leads to an emotion ("\textit{She gave me a gift}"). Most annotation schemes have implicitly focused on the \textit{labeled} emotion, overlooking the other two expression modes. Recently, annotation schemes based on appraisal theories implicitly concern themselves with the \textit{suggested} emotion \cite{troiano-etal-2023-dimensional}. Micheli's theory thus analyzes the different types of verbal signs humans use to infer expressed emotions. In a complementary manner, theories of cognitive pragmatics are interested in the psychological mechanisms used to infer what is communicated, especially the emotions expressed by these different types of signs. In the next section, we will hypothesize that the sign categories distinguished by Micheli correspond to different sources of inferences postulated by cognitive pragmatics.

\section{Cognitive Pragmatics and Emotional Communication\label{modele_detective}}


\paragraph {Two analyses of communication.} Cognitive pragmatics is the branch of cognitive science concerned with how agents use and interpret signs in communication. In this and related branches, it is common to distinguish between two broad ways to analyze communication: the "dictionary analysis" (a.k.a. the "code", "semiotic", or "semantic" model) and the "detective analysis" (a.k.a. the "Gricean", "inferential", or "pragmatic" model) \cite{sperber_relevance_1995,aloni_semantics-pragmatics_2016,heintz_expression_2023}.

\paragraph {Dictionary analysis.} The dictionary analysis depicts communication as a sender who intentionally or unintentionally encodes information into a signal that the receiver decodes. Vitally, prior to the communicative exchange, the sender and the receiver must share the same code. A code here is understood as a pre-established pairing between kinds of stimuli (symbolized by "<…>") and sets of information (symbolized by "[…]"). For instance, the Morse code consists of a pairing between <combinations of short and long signals> and [letters] that senders and receivers must share to communicate with it. Codes can be conventional, as the Morse code is and as is the formal semantics of a language: a code made of syntactical and lexical rules that pairs <strings of words> with [sentential meanings] \cite{heim_semantics_1998}. Codes can also be non-conventional or "natural" \cite{wharton_natural_2003,bonard_natural_2023}. For instance, bees are thought to use a code pairing their <dances> with the [location of nectar]. As mentioned in section \ref{theorie_emotion}, humans are thought to use a code pairing types of <facial expressions> with types of [emotions expressed].

The main limitation of the dictionary analysis is that codes sometimes \textit{underdetermine meaning}: The pre-established pairings between <types of stimuli> and [sets of information] are sometimes insufficient to account for the information communicated. Paradigmatically, in \textit{conversational implicatures} \cite{grice_logic_1975}, the utterer implicitly communicates information beyond what is linguistically encoded, beyond what is determined by syntactical and lexical rules. For instance \cite{wilson_relevance_2006}, if Peter asks, "Did John pay back the money he owed you?" and Mary answers, "He forgot to go to the bank.", Peter will readily understand that Mary means "no". However, the relevant code – the rules pairing <English grammar and lexicon> with [sentential meaning] – is insufficient to account for this since the code only tells you that John forgot to go to the bank.

Codes underdetermine the meaning of verbal expressions of emotions as well. To illustrate, let us go back to Micheli’s typology: \textit{labeled}, \textit{displayed}, and \textit{suggested} emotions \cite{micheli2013}. As far as \textit{labeled} emotions are concerned, the dictionary analysis does quite well thanks to the pairing between <emotion words> (\textit{e.g.}, happy, amazing, sadly) with the [emotion kinds] they refer to. However, even \textit{labeled} emotions sometimes do not encode all that is communicated. For instance, "I am happy now" is explicit about the kind of emotion expressed but does not encode what the emotion is about. Nevertheless, we often correctly infer such information in the relevant context. The dictionary analysis fairs even less well with \textit{displayed} emotions because these are often ambiguous. For instance, interjections such as "Wow!", "Damn!", "Fuck!", "Shit!", "Ah!", and "Oh!" though they readily display that the utterer undergoes an emotion, can express various positive and negative emotions. Furthermore, these interjections don’t encode what emotions are about. However, receivers usually correctly infer these pieces of information. The dictionary analysis regarding \textit{suggested} emotions is even more limited. Depending on what the person expressing their emotion believes or desires, a phrase that only suggests emotions can communicate pretty much any kind of emotion. Imagine, for instance, that someone says, "The ship has black sails.". In a certain context, this apparently vapid sentence may poignantly convey intense emotion – because, say, it means that the son of the utterer died, as in the story of Aegeus and Theseus. Note that, beyond verbal expression, most, if not all, types of emotional expressions also underdetermine what emotions are expressed. Facial expressions or acoustic cues (\textit{e.g.}, screams, laughter, sighs) also communicate different emotions given different contexts \cite{aviezer_angry_2008,teigen_is_2008,vlemincx_why_2009,barrett_context_2011,barrett_emotional_2019,bonard_underdeterminacy_2023}. The dictionary analysis is thus also insufficient for these kinds of emotional expressions. 

So, how do humans disambiguate emotional expressions in cases where codes underdetermine what is communicated? If we trust contemporary cognitive pragmatics, the answer should be found in the detective analysis of communication.

\paragraph{Detective analysis.} What we call "the detective analysis" is constituted by a family of theories developed by Paul Grice \cite{grice_meaning_1957,grice_studies_1989} and his heirs (for reviews, see \citet{bonard_meaning_2021}, chapter one and appendix). Note that although our presentation aims to remain balanced, no universally accepted version of this analysis exists.

As mentioned, the detective analysis was developed to account for conversational implicatures, cases where what is communicated goes beyond what is conveyed through conventional meaning, as in Peter and Mary’s example above. To do so, the detective analysis conceptualizes linguistic interpretation as a type of abductive reasoning – \textit{i.e.}, as an inference that seeks the simplest and most likely conclusion given the evidence available. The analysis spells out three main sources of evidence:
\begin{enumerate}
    \item\textit{Codes}, \textit{i.e.}, pre-established pairings between types of stimuli and sets of information, \textit{e.g.}, English syntactical and lexical rules; the codes for verbal and nonverbal emotional expressions. As we saw, expressions using labeled (\textit{e.g.}, "I'm happy") and displayed emotions (\textit{e.g.}, "Damn!") are partially understood through such codes, though they are too ambiguous to account for all that is communicated.
    \item\textit{Pragmatic expectations}, \textit{i.e.}, how people are expected to behave in given contexts, particularly the kind of signal they receive. For instance, in conversations, people are expected to say things relevant to the question under discussion (see \citet{grice_logic_1975}’s maxims of conversation). For this reason, although what is literally encoded in Mary's reply is that John forgot to go to the bank, Peter will nevertheless expect this to be relevant to the question he asked. Similarly, we expect someone’s emotional expressions to be about something relevant to their concerns \cite{wharton_relevance_2021,bonard_beyond_2022}. For instance, if someone says "Damn!" after receiving a surprisingly nice compliment, we expect the compliment to be particularly relevant to the person and will interpret the interjection accordingly.
    \item\textit{Common ground}, \textit{i.e.}, the information presumed to be shared by the participants in the exchange \cite{stalnaker_common_2002}. For instance, Mary and Peter both presume that a bank is a place where one can withdraw money. Similarly, we usually presume that receiving a compliment is something that one seeks, especially if it is surprisingly nice – though this is not always part of the common ground, \textit{e.g.}, if the complimenter is the complimentee's archenemy. The common ground also allows us to understand that Aegeus can express deep despair with the sentence <<~The ship has black sails.~>>.
\end{enumerate}

 Based on these three sources of evidence, the detective analysis further postulates that the interpreter uses \textit{mindreading} abilities (\textit{i.e.}, theory of mind, mentalizing, or social cognition) to infer what is the most likely piece of information that is implicitly communicated – \textit{e.g.}, Peter infers that Mary meant "no" and we infer that the person saying "Damn!" is probably pleased. Finally, the detective analysis specifies that the information so inferred is added to the common ground shared by participants in the exchange so that it may be a new source of evidence in the upcoming exchanges.

Let us note that the detective analysis predicts that the ability to correctly infer what is communicated by emotional expressions heavily depends on one's mind-reading capacities. Corroborating this prediction, children or people on the autistic spectrum may struggle to infer implicit meaning correctly, \textit{e.g.}, conversational implicatures \cite{ball_conversational_2024} or in expressions using suggested emotions \cite{blanc_production_2017}. 

\section{Research Directions for Emotion Analysis\label{directions}}

\subsection{Towards a Unified Annotation Scheme}


Training models on data annotated with a scheme that reflects the multifaceted nature of emotions is desirable to improve the capacity of language models to understand emotions. Such a scheme would need to integrate different perspectives on the emotional phenomena to allow for better study comparisons. This would also increase the performance and generalization of models.

\paragraph{Attempts at unification.} Several recent studies attempt to unify different ways of annotating emotion in text. \citet{campagnano-etal-2022-srl4e} propose a new annotation scheme that unifies various schemes on emotion semantic roles. To choose a set of shared categories, the different discrete emotions from the schemes were converted to the basic emotions of Plutchik's theory \cite{plutchikNatureEmotionsHuman2001}. \citet{klingerWhereAreWe2023} explores the divergences and commonalities between semantic role labeling of emotions and approaches based on appraisal theories. The study identifies several research directions, such as using appraisal variables to improve the task of detecting emotion causes, or analyzing experiencer-specific appraisals \cite{weggeExperiencerSpecificEmotionAppraisal2023}. These studies show that combining schemes allows knowledge transfer between tasks, increasing performance and generalization.


\paragraph{In search of a common framework.} What we have previously referred to as "the integrated framework for emotion theories" (section \ref{theorie_emotion}) aims to reconcile the main emotion theories in psychology \cite{schererTheoryConvergenceEmotion2022}. In our view, it represents a strong candidate to provide a common framework for annotation schemes. As mentioned in section \ref{theorie_emotion}, this model considers that emotion consists of synchronized changes in different components: the appraisal process, action tendencies, bodily changes (motor expressions and physiological responses), and subjective feelings. Research in emotion analysis must draw from the recent debates in the psychology of emotions to bring existing annotation schemes into dialogue on a solid theoretical basis and, ideally, construct a unified annotation scheme.

\paragraph{Emotion comprises several interacting components.} A unified annotation scheme could clarify some gray areas in emotion analysis, such as the lack of clear definitions for emotion semantic roles (\textit{e.g.}, experiencer, cause, and target). It could also better situate existing schemes. For example, annotating discrete emotions and affective dimensions emphasizes subjective feeling, whereas annotating cognitive dimensions emphasizes appraisals. Few schemes account for physiological responses, motor expressions, and action tendencies. More generally, few schemes consider all components. \citet{kim-klinger-2019-analysis} analyze the communication of emotions in fiction through descriptions of subjective sensations, postures, facial expressions, and spatial relations between characters. \citet{casel-etal-2021-emotion} associate text spans with categories corresponding to Scherer's emotional components. \citet{cortal:hal-03805702,cortal-etal-2023-emotion} structure emotional narratives according to components similar to Scherer's. Each text span corresponds to observable behaviors, thoughts, physical feelings, or appraisals. To our knowledge, no annotation schemes attempt to capture the interaction between components. Generally, emotion analysis pays little attention to the dynamic nature of emotion and the synchronization of its various components.

\paragraph{Improving the clarity of annotation guides.} We note that few studies psychologically justify the choice of different objects to detect in the text. Emotion analysis needs to develop a systematic approach to compare annotation guides with one another, thereby precisely understanding how different annotation schemes capture emotion. Thus, these schemes must draw from psychological theories (section \ref{theorie_emotion}) but also from linguistic theories (sections \ref{limitations} and \ref{modele_detective}) to identify linguistic markers that verbalize emotion. With clear annotation guides, it would be easier for research teams to focus on points of convergence between schemes.

\subsection{Better Knowledge Use and Environmental Interaction}

In natural language processing, \textit{prompting} refers to supplying a tailored input to a language model, aiming to direct its generation process towards a desired response \cite{brownLanguageModelsAre2020a}. Numerous prompting methods draw inspiration from human cognition to improve the performance of language models \cite{zhangIgnitingLanguageIntelligence2023}. These methods propose generating reasoning steps \cite{weiChainofThoughtPromptingElicits2023,kojimaLargeLanguageModels2023}, reasoning through multiple generated responses \cite{wangSelfConsistencyImprovesChain2023,yoranAnsweringQuestionsMetaReasoning2023}, facilitating communication by rephrasing questions \cite{dengRephraseRespondLet2023c}, and self-improving with its own generated feedback \cite{madaanSelfRefineIterativeRefinement2023,yuanSelfRewardingLanguageModels2024}.

\paragraph{Prompting methods for emotional understanding.} Most methods have been explored to improve model performance on tasks requiring formal reasoning \cite{zhangIgnitingLanguageIntelligence2023}. We believe it is possible to adapt these methods or even create new ones to improve model performance on tasks requiring social reasoning, such as emotional understanding. It would be interesting to rely on the ability of language models to act as character simulators \cite{shanahanRolePlayLarge2023,luLargeLanguageModels2024}, capable of adopting multiple perspectives to change style \cite{deshpandeToxicityChatGPTAnalyzing2023}, solve tasks requiring expert knowledge \cite{xuExpertPromptingInstructingLarge2023}, or simulate discussions to encourage exploration \cite{wangUnleashingCognitiveSynergy2023,liangEncouragingDivergentThinking2023}. \citet{zhouHowFaRAre2023} enhance the ability of language models to make relevant inferences for solving theory of mind tasks. They propose a reasoning structure that anticipates future challenges and reasons about potential actions. More globally, a major challenge in natural language processing is finding suitable reasoning structures to effectively use the internal knowledge of models \cite{kojimaLargeLanguageModels2023,zhouHowFaRAre2023,zhouSelfDiscoverLargeLanguage2024}. The contribution of the detective analysis (section \ref{modele_detective}) could prove valuable here: prompts that explicitly ask models to seek evidence from the three sources highlighted by this analysis could lead to better performance and explainability. Finally, the integrated framework for emotion theories (section \ref{analyse_des_emotions}) can serve as inspiration for prompts that aim to exploit all the different facets of emotions rather than focusing on just one of them (\textit{e.g.}, subjective feeling).

\paragraph{Interaction with the environment.} Current language models, trained solely on predicting missing words, have essentially mastered linguistic codes, \textit{i.e.}, lexical and syntactic rules (section \ref{modele_detective}), which \citet{mahowaldDissociatingLanguageThought2023a} call "formal linguistic competence". However, they struggle to perform well on tasks relying on what \citet{mahowaldDissociatingLanguageThought2023a} call "functional linguistic competence", \textit{i.e.} the skills required to use language in real-world situations. These skills centrally involve the mechanisms postulated by the detective analysis – in particular, sharing a common ground and having sensible pragmatic expectations (section \ref{modele_detective}). To address this limitation, studies augment language models with external modules like a mathematical calculator \cite{schickToolformerLanguageModels2023}, a web browser \cite{gurRealWorldWebAgentPlanning2023}, or a virtual environment \cite{parkGenerativeAgentsInteractive2023a}. Through tool manipulation, language models intertwine reasoning with action and can thus effectively combine internal with external knowledge \cite{yaoReActSynergizingReasoning2023}. This point is crucial to develop models that exhibit human-like social behaviors. For example, \citet{parkGenerativeAgentsInteractive2023a} show that observation, planning, and reflection are important components for increasing the credibility of behaviors in a virtual environment. Research on human communication can help highlight relevant abilities to augment language models (\textit{e.g.}, with external modules). This surely applies to emotional communication as well: models could be complemented with modules encapsulating, for instance, our knowledge of codes for emotional expressions, of how kinds of appraisals relate to kinds of emotions, and of how we expect people undergoing emotions to behave, along the lines sketched in sections \ref{theorie_emotion} and \ref{modele_detective} above.

\subsection{Better Benchmarks for Emotional Understanding} 

Recent benchmarks evaluate language models on specific aspects of emotional understanding \cite{wangEmotionalIntelligenceLarge2023,paechEQBenchEmotionalIntelligence2024}, but they don't consider its full richness \cite{scherer_componential_2007,mayer_human_2008,oconnor_measurement_2019}. For example, \citet{paechEQBenchEmotionalIntelligence2024} assesses emotional understanding by predicting the intensity of multiple emotions in conflict scenes. Some benchmarks evaluate models on related tasks, such as sentiment analysis \cite{zhangSentimentAnalysisEra2023} and theory of mind \cite{zhouHowFaRAre2023,maHolisticLandscapeSituated2023,kimFANToMBenchmarkStresstesting2023,gandhiUnderstandingSocialReasoning2023}. However, no benchmark specifically proposes to evaluate the multiple facets of emotions that affective sciences reveal (section \ref{theorie_emotion}). Therefore, it is difficult to know whether current models are efficient for emotional understanding.

This limitation is compounded by the fact that it is difficult to clearly determine which properties of emotional understanding are to be evaluated. We believe that evaluating language models should be grounded in research on human emotional communication, especially psycholinguistics. For example, before the age of ten, basic emotions (e.g., joy or sadness) are better remembered than complex emotions (e.g., pride or guilt) \cite{davidsonChildrenRecallEmotional2001,creissenQuelleRepresentationDifferentes2017a}. From six to ten years old, \textit{labeled} emotions are better understood than \textit{suggested} emotions \cite{blancComprehensionContesEntre2010,creissenQuelleRepresentationDifferentes2017a}. Another example of relevant studies concerns the difficulty that autistic people have in understanding different types of emotional expressions \cite{ball_conversational_2024}. These studies suggest that, for humans, different types of emotions and different modes of emotional expression are more or less difficult to interpret. It would be desirable for benchmarks to evaluate language models in ways that reflect the relative difficulty of tasks for humans. Such a project would certainly benefit from research in cognitive pragmatics (section \ref{modele_detective}), knowing, for example, that people with communication disorders have difficulty understanding conversational implicatures \cite{ball_conversational_2024}, which indicates that the different sources of evidence distinguished by the detective analysis are associated with different levels of difficulty.

We believe the concept of emotion should be addressed through its relationship with text understanding, i.e., the ability of a reader to construct a mental representation of a situation in a text \cite{zwaanSituationModelsLanguage1998}. Thus, we would need to go beyond current conceptualizations of emotion in natural language processing (section \ref{comment_emotion_annotee}) to consider the diversity of linguistic markers used to verbalize emotion (section \ref{limitations}) as well as the different types of emotion (basic or complex) from psycholinguistic research (section \ref{theorie_emotion}). Inspired by previous studies, \citet{etienne-etal-2022-psycho} propose an annotation scheme that considers emotion expression modes and types of emotion. Future benchmarks assessing the ability of language models to analyze emotions should consider such annotation schemes, which, as we have recommended, seek to be solidly based on relevant research in cognitive science.

\section{Conclusion}

Emotion analysis has several limitations that, we believe, are partially due to a lack of communication with other disciplines and, in particular, cognitive science. We propose exploiting cognitive science research on emotions and communication to address some limitations, especially what we called "the integrated framework" in emotion theories and "the detective analysis" in cognitive pragmatics. We suggest that this opens the way for constructing new annotation schemes, methods, and benchmarks for emotional understanding that consider the multiple facets of human emotion and communication.

\section*{Limitations}

We propose a theoretical perspective on emotion analysis in natural language processing. We believe it would benefit the emotion analysis community to adopt an interdisciplinary approach by drawing from cognitive science theories to address certain existing limitations in the research field. In practice, this is a challenging task. Although we focus on concrete actions that could be undertaken soon (for example, clarifying annotation guidelines), we recognize that our contribution involves speculative research directions. In future research, it would be desirable to complement these speculative aspects with more concrete proposals, notably with empirically testable hypotheses and implementable algorithms.



\section*{Ethics Statement}

We have not conducted any experimentation or published any data or models in this paper. The present research aims to better understand human emotional communication, not to develop tools for automatically detecting individuals' private subjective states. While we believe our paper does not present direct ethical concerns, the research directions it raises could indirectly harm individuals and societal structures. Although we have highlighted the potential benefits of natural language processing applications (such as emotion regulation tools), it is crucial to ensure that the development and use of such tools do not have any adverse effects in the future.




\section*{Acknowledgements}

Thanks to the AI-PHI group\footnote{\url{https://ai-phi.github.io/}} for making our collaboration possible and for insightful discussions. Thanks also to the ACL ARR 2024 February reviewers for useful and constructive feedback.
\bibliography{custom,anthology,eacl_2023_emotion_regulation,zotero_gustave,constant}




\end{document}